# AI for Maritime Security: Comparative Evaluation of CNN and Vision Transformer Architectures for Maritime Object Detection

Dr. Ismet Gocer *
Southampton Solent University
School of Technology and Maritime Industries
Southampton, United Kingdom
ismet.gocer@solent.ac.uk
* Corresponding Author

Prof. Zakirul Bhuiyan
Southampton Solent University
School of Technology and Maritime Industries
Southampton, United Kingdom
zakirul.bhuiyan@solent.ac.uk

Dr. Shakeel Ahmad
Southampton Solent University
School of Technology and Maritime Industries
Southampton, United Kingdom
shakeel.ahmad@solent.ac.uk

Dr. Raza Hasan
Southampton Solent University
School of Technology and Maritime Industries
Southampton, United Kingdom
raza.hasan@solent.ac.uk

**Funding Statement:** This paper is produced based on research conducted under the SEAGUARD project, a European Union Horizon Europe research and innovation project (Grant Agreement No. 101168489).

**Abstract**

This study aims to enhance maritime security by using advanced Artificial Intelligence (AI) and Computer Vision (CV) techniques. For this purpose, it was designed and assessed intelligent object detection systems that can detect the presence of ships on the sea surface under different real-time environments. To achieve this goal, a maritime image dataset with 6,468 images was used, covering different weather conditions like cloudy, foggy, rainy, and sunny environments.

Six deep learning architectures were evaluated, including a base Convolutional Neural Network (CNN) model, four transfer learning models (Xception, VGG16, MobileNetV2, and EfficientNetV2L), and a Vision Transformer (ViT) model. The models were compared using multiple performance indicators, including accuracy, Type I and Type II errors, model size, and video processing time.

The results show that model performance varies depending on computational constraints and deployment conditions. While lightweight architectures are suitable for resource-limited devices, the ViT achieved the best overall performance, reaching 100% accuracy with the lowest error rates and the fastest video processing time. The findings highlight the potential of AI-driven computer vision systems for maritime surveillance, border protection, and autonomous navigation.

**Graphical Abstract**

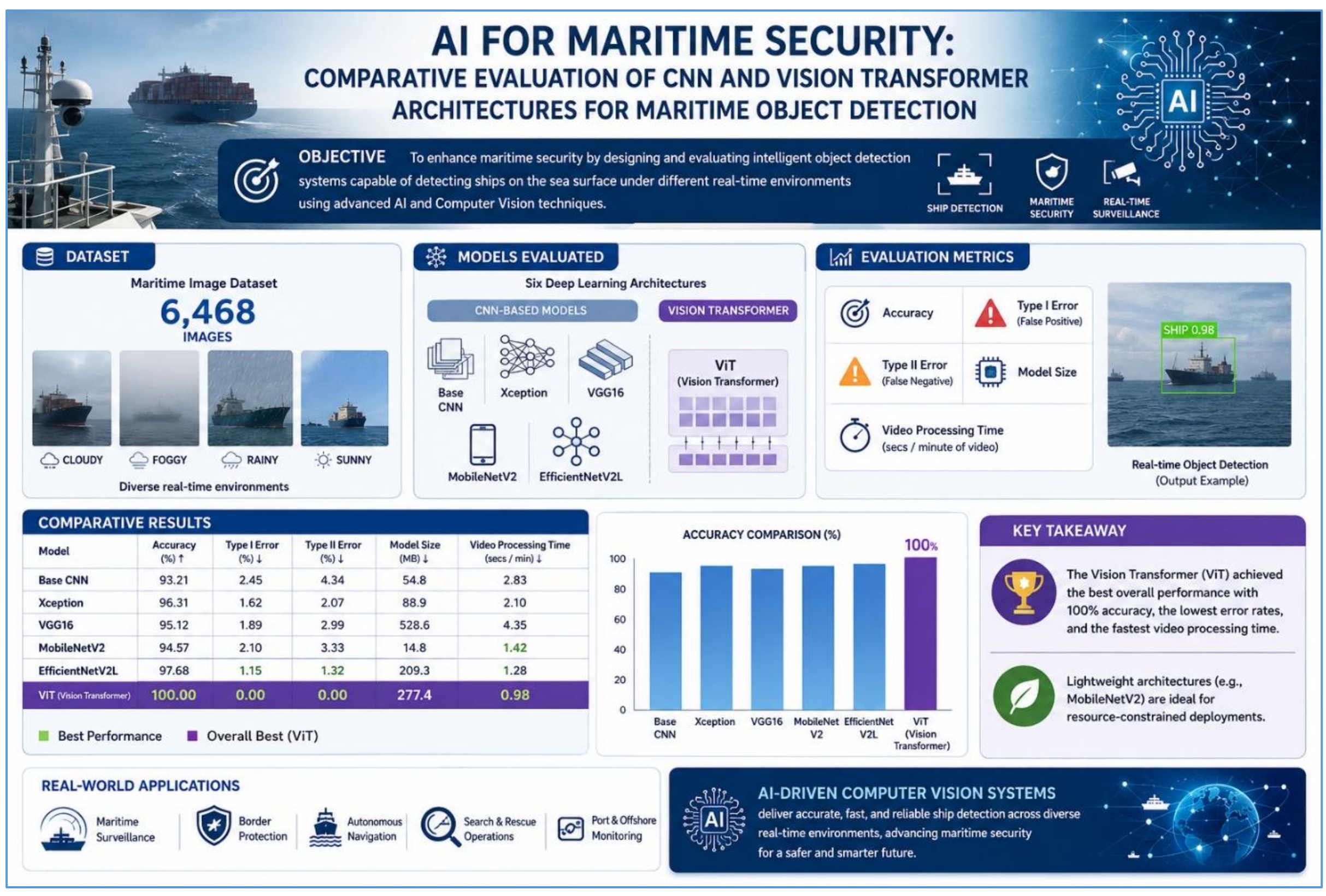

## 1. Introduction

Recent years have witnessed a dramatic escalation in threats to the maritime domain, including illicit trafficking, unregulated fishing, environmental violations, and covert vessel operations. More than 80% of global trade is carried by sea, and undersea infrastructure supports nearly all international data traffic, underscoring the strategic importance of maritime security. Deep-learning-based Artificial Intelligence (AI) and Computer Vision (CV) techniques are emerging as critical enablers of this domain awareness, able to process vast multimodal data streams in real time and to flag anomalous behaviour or objects of interest that elude conventional systems (Munim, 2020). Unlike traditional base surveillance relying on radar sweeps, human patrols, and satellite overviews, AI-based CV systems can detect, classify, and track maritime targets such as small boats or floating debris with high accuracy, even under challenging sea states and lighting conditions (Talpur et al., 2025). These capabilities enhance operational response and situational awareness for coast guards and naval forces, reduce human workload, improve monitoring coverage, and accelerate decision-making in dynamic environments.

Building upon these advancements, AI and computer vision now function as pivotal technologies for strengthening maritime security and operational awareness (Talpur et al., 2025). They enable intelligent detection and recognition of vessels, identification of abnormal activities, and continuous monitoring of maritime environments through visual and sensor data (Liu et al., 2022). By automating the processing of heterogeneous data streams, including optical imagery, SAR, and AIS, AI-based systems provide early-warning capabilities and improve decision-making accuracy in complex sea states (Karst et al., 2025). Although deep learning-based vision models have demonstrated strong performance in maritime surveillance applications, supporting coastal agencies in detecting illegal fishing, smuggling, and unauthorised vessel movement, several challenges remain, such as limited annotated data, domain adaptation difficulties, and deployment on resource-constrained platforms (Talpur et al., 2025). These advancements show that AI-enhanced vision systems not only extend surveillance reach but also enable proactive threat detection by shifting from passive observation to automated, real-time decision making. Nevertheless, researchers have noted that data scarcity, domain adaptation, and real-time integration continue to constrain large-scale deployment in practical maritime operations (Juan et al., 2025). Addressing these challenges demands advanced architectures such as Transfer Learning and Vision Transformers (ViT) can extract both global and local features (Shedrawi et al., 2024), thereby improving detection accuracy and robustness in real-world maritime scenarios. In this context, this study aims to develop new sea surveillance tools and compare the performance of base CNN, transfer learning algorithms and ViT for object recognition in the maritime domain.

In this context, this study aims to demonstrate the applicability of five AI-based computer vision models and Vision Transformer (ViT), that have already proven their effectiveness in previous experimental research. These models are used as complementary tools for different tasks in maritime vision systems. ViT is employed for object detection due to its strong capability in capturing global visual representations. The study presents the implementation of these approaches and reports their results across specific use cases to indicate their suitability for varying operational needs in maritime surveillance. In addition, the study discusses the challenges encountered during the implementation of these methods and the practical solutions developed by the authors, aiming to provide valuable insights for future researchers and industry professionals.

The main contributions of this study are summarized as follows:

i. **Systematic comparison of multiple computer vision architectures for maritime object detection.**
This study presents a comprehensive evaluation of six deep learning models, including a base CNN architecture, four transfer learning models (Xception, VGG16, MobileNetV2, and EfficientNetV2L), and a transformer-based model (Vision Transformer). The comparison is conducted using multiple performance indicators, including training and testing accuracy, Type I and Type II errors, model size, and computational processing time.

ii. **Detailed methodological framework for data preprocessing in transfer learning applications.**
Data preprocessing is a critical stage in transfer learning-based computer vision systems and is often implemented incorrectly or insufficiently described in many studies. This paper provides a structured and detailed explanation of how preprocessing pipelines should be applied across different deep learning architectures. By clarifying these procedures, the study contributes to improved methodological transparency, reproducibility, and reliability in transfer learning applications.

iii. **Scenario-based evaluation of computer vision models for real-world deployment.** Rather than focusing solely on predictive accuracy, this study evaluates model performance from a practical deployment perspective by jointly analysing model size, computational requirements, processing speed, and error types. This analysis demonstrates how different models may be more suitable under different operational conditions, such as resource-constrained edge devices or high-capacity maritime monitoring systems.

iv. **Practical guidelines for model selection in maritime monitoring systems.** Based on the experimental results, the study provides practical insights into selecting appropriate computer vision models for maritime surveillance applications. The findings highlight that lightweight models may be preferable in environments with limited computational resources, while transformer-based architectures such as Vision Transformer offer superior performance in systems where detection accuracy and reliability are critical.

The remainder of this paper is organized as follows. Section 2 presents the literature review related to maritime object detection and recent developments in computer vision architecture. Section 3 describes the methodology used in this study. Section 4 introduces experimental design and provides a comparative evaluation of the applied models. Section 5 discusses the main findings and their implications for maritime monitoring systems. Section 6 presents the conclusions of the study. Finally, Section 7 outlines potential directions for future research.

## 2. Literature Review for Computer Vision Techniques in Maritime Security

The application of computer vision techniques to maritime security began with pioneering studies in the mid-2010s, when researchers first explored camera-based monitoring for sea-surface object detection and tracking. Early contributions such as those by Schwendeman and Thomson (2015), Kristan et al. (2016), and Prasad et al. (2017) investigated video-based maritime surveillance using electro-optical sensors, establishing baseline algorithms for horizon detection, motion compensation, and background subtraction specifically adapted to dynamic maritime environments and capable of handling the instability of sea platforms and illumination variations. These works demonstrated the feasibility of automated vessel monitoring but relied heavily on handcrafted features and controlled datasets, indicating the technical limitations of early vision systems. Together, these studies demonstrated the feasibility of applying computer vision methods to maritime monitoring and motivated the development of open datasets for reproducible research. They also showed that traditional vision algorithms, including optical flow, histogram of oriented gradients (HOG), and background modelling, could support vessel detection under challenging visual conditions and varying sea states.

As publicly available datasets expanded, research interest in image-based maritime perception accelerated. The introduction of the Singapore Maritime Dataset (SMD) and the SeaShips dataset (Shao et al., 2018) marked a decisive turning point by enabling reproducible benchmarking of ship detection and tracking algorithms. This data availability substantially improved research reproducibility and allowed comparative evaluation across algorithms. Building upon this foundation, Ma et al. (2018) and Fan et al. (2019) demonstrated that convolutional neural networks (CNNs) significantly outperformed handcrafted feature-based methods, signalling the transition from traditional computer vision to deep learning in maritime applications. CNN-based architectures achieved superior generalisation and detection accuracy in complex maritime scenes (Wang et al., 2021; Ribeiro et al., 2022), although they often struggled with small or distant targets and were sensitive to environmental noise such as reflections and low contrast.

Following these developments, Wang et al. (2021) and Ribeiro et al. (2022) emphasised the growing importance of deep learning architectures for achieving robust ship detection under challenging maritime conditions, including waves, occlusions, reflections, and low contrast. Their findings highlighted that conventional handcrafted pipelines could not adequately capture the complex variability of sea environments, thereby reinforcing the need for more diverse and balanced datasets to enhance model generalisation. In response to these limitations, convolutional neural network (CNN)–based frameworks such as Faster R-CNN (Ren et al., 2015) and SSD (Liu et al., 2016) became the dominant paradigms in ship detection and classification, providing significant improvements in accuracy and adaptability across varied maritime scenes.

A new phase of progress emerged with the advent of Transformer-based architectures and end-to-end learning frameworks. Carion et al. (2020) introduced the Detection Transformer (DETR), demonstrating that attention mechanisms could effectively capture long-range contextual dependencies, a critical factor in distinguishing vessels from cluttered sea backgrounds. To overcome CNN limitations, Transformer-based models introduced global context modelling and end-to-end detection pipelines, offering improved robustness to cluttered visual environments and reducing dependency on handcrafted priors. Building upon this, Li et al. (2022) proposed ViTDet, showing that Vision Transformers (ViT) achieve competitive accuracy with minimal architectural modification, producing more stable and context-aware representations in complex maritime imagery.

Several authors, including Carion et al. (2020), Khan et al. (2022), and Li et al. (2022), have argued that Transformer-based models such as ViT offer superior global feature extraction and robustness under cluttered visual conditions. Collectively, these studies reveal a clear methodological evolution from handcrafted pipelines toward hybrid deep architectures and position Vision Transformers as complementary state-of-the-art solutions, combining context-rich perception with real-time operational efficiency.

## 3. Methodology

In this study six algorithms were used to object detection on sea surface. Which is a base CNN algorithm, four transfer learning algorithms (XCeption, VGG16, MobileNetV2 and EfficientNetV2L) and Vision Transformer (ViT). We can take a look that models:

### 3.1. Baseline Convolutional Neural Network

As a baseline model, we implemented a manually designed Convolutional Neural Network (CNN) architecture to establish a controlled reference point prior to evaluating modern detection architectures. Convolutional neural networks have been widely adopted in visual recognition tasks due to their ability to automatically learn hierarchical spatial features through convolutional operations and parameter sharing (LeCun et al., 1998; Krizhevsky et al., 2012).

Unlike pre-defined architectures, the CNN in this study was constructed manually, allowing full architectural flexibility. This approach enabled systematic control over model depth, convolutional filter sizes, number of feature maps, and fully connected layers. In particular, the manual design allowed us to:

- Adjust the number of hidden layers to analyse depth-related performance variations.

- Integrate Batch Normalization to stabilize training and accelerate convergence (Ioffe and Szegedy, 2015).
- Incorporate Dropout layers to mitigate overfitting, particularly important given the heterogeneous nature of maritime imagery (Srivastava et al., 2014).
- Experiment with different activation functions and kernel configurations.

This flexible configuration facilitated the development of multiple CNN variants under identical training protocols. By modifying architectural components incrementally, we were able to analyse how structural complexity influenced detection performance across datasets with varying distributional characteristics.

The base CNN therefore served two key purposes:

- It provided a transparent and interpretable baseline model.
- It enabled controlled experimentation to assess the impact of architectural depth and regularization strategies on maritime object detection.

Such modular CNN construction remains a valuable methodological approach when investigating domain-specific performance behaviour, particularly in heterogeneous maritime environments.

### 3.2. Transfer Learning

Although manual (base) CNN architectures are highly flexible, determining the optimal image size, number of hidden layers, filter configurations, and the inclusion of components such as dropout or batch normalization typically requires extensive experimentation. Identifying a well-performing configuration often involves iterative tuning and repeated trials, which can be computationally demanding and time-consuming.

To mitigate this complexity, Transfer Learning (TL) approaches can be employed. In practice, TL is implemented using pre-trained models, which are built upon empirically validated architectures and trained on large-scale datasets to learn robust feature representations. The researcher selects a suitable pre-trained model according to the dataset characteristics and available computational resources and transfers its learned representations and pretrained weights to the target task. Thus, Transfer Learning is fundamentally based on the principle of reusing previously acquired knowledge for a related problem domain.

The complete list of these pre-trained models is available at the Keras Applications repository[1]. The models provided in this collection vary in terms of architectural depth, computational complexity, and parameter size, allowing researchers to select an appropriate approach based on their hardware capacity and specific task requirements. In this study, four representative architectures with differing structural characteristics and model capacities, namely Xception, VGG16, MobileNetV2, and EfficientNetV2L, were selected from this list and implemented.

The typical application domains of these selected architectures are discussed in the following section. In addition to their architectural differences, an important methodological consideration must also be addressed. Each architecture requires model-specific preprocessing steps, and following these procedures is essential for the proper use of Transfer Learning. Pre-trained models are optimized under specific input normalization schemes during large-scale training, particularly on ImageNet (Krizhevsky et al., 2012; Simonyan and Zisserman, 2015). Since Transfer Learning assumes consistency between training and target data distributions (Weiss et al., 2016; Tan and Le, 2019), deviations from the original preprocessing strategy may lead to performance degradation. Therefore, the required preprocessing protocols are described alongside each selected model. All pre-trained Transfer Learning models used in this study necessitate architecture-specific input preprocessing aligned with their original ImageNet training configurations. The corresponding preprocessing strategies implemented for each model are provided in Section 4.

[1] https://keras.io/api/applications/

### 3.2.1. XCeption

Xception (Extreme Inception) is a deep convolutional neural network architecture introduced by François Chollet in 2017 (Chollet, 2017). It can be considered an extension of the Inception family, where standard Inception modules are replaced with depthwise separable convolutions. Unlike traditional convolutions that jointly learn spatial and channel-wise correlations, Xception separates these operations into two steps: spatial filtering is applied independently to each channel, followed by pointwise convolution to combine channel information. This design significantly reduces the number of parameters while maintaining strong representational capacity.

A key advantage of Xception is its computational efficiency. Depthwise separable convolutions lower memory usage and computational cost compared to conventional convolutional layers. Empirical results show that Xception achieves superior performance compared to InceptionV3 on large-scale datasets such as ImageNet (Chollet, 2017). For this reason, it has become a widely adopted architecture in Transfer Learning applications for image classification and detection tasks.

However, compared to lightweight architectures such as MobileNet, Xception requires relatively higher computational resources. When applied to smaller datasets, careful adaptation is necessary to avoid performance degradation. In this study, Xception was selected due to its balance between efficiency and high representational power. Its ability to extract detailed spatial features makes it suitable for maritime object detection scenarios, where accurate visual pattern recognition is essential.

### 3.2.2. VGG16

VGG16 is a deep convolutional neural network architecture proposed by Simonyan and Zisserman (2015) as part of the Visual Geometry Group (VGG) at the University of Oxford. It was introduced in the ImageNet Large Scale Visual Recognition Challenge (ILSVRC) and became widely recognized for demonstrating that increasing network depth using small 3×3 convolutional filters significantly improve classification performance.

The VGG family includes several variants, notably VGG16 and VGG19, which differ primarily in depth. VGG19 contains three additional convolutional layers compared to VGG16, potentially allowing it to learn more complex hierarchical feature representations. In scenarios involving highly intricate visual patterns or very large-scale datasets, VGG19 may offer marginal performance improvements. However, the increased depth also results in higher computational cost and a greater number of parameters. For this reason, VGG16 is more commonly preferred in practical applications, as it provides a strong balance between representational capacity and computational efficiency (Simonyan & Zisserman, 2015).

Due to its structural simplicity, stable training behaviour, and well-established performance, VGG16 remains a widely adopted baseline architecture in Transfer Learning studies. In this work, VGG16 was selected as a classical deep CNN benchmark to enable comparison with more computationally optimized architectures within the maritime object detection framework.

### 3.2.3. MobileNetV2

MobileNet is a family of lightweight convolutional neural network architectures developed for efficient deep learning on mobile and embedded devices (Howard et al., 2017; Sandler et al., 2018). The series includes several versions, notably MobileNet, MobileNetV2, and MobileNetV3, each introducing architectural refinements to improve efficiency and accuracy under constrained computational settings.

MobileNetV2, proposed by Sandler et al. (2018), introduces two key design concepts: inverted residual blocks and linear bottlenecks. Unlike traditional residual connections, inverted residuals connect low-dimensional representations, reducing computational cost while preserving essential feature information. Depthwise separable convolutions are also employed to significantly decrease the number of parameters and floating-point operations. This architecture achieves a favourable trade-off between model size, speed, and classification performance.

Compared to deeper and more parameter-heavy models such as VGG or Xception, MobileNetV2 is particularly suitable for scenarios where computational resources are limited or real-time inference is required. While MobileNetV3 further enhances efficiency through neural architecture search and additional optimization strategies (Howard et al., 2019), MobileNetV2 remains widely adopted due to its structural clarity, stable performance, and broad framework support.

### 3.2.4. EfficientNetV2L

Recent advances in convolutional neural network design have emphasized not only depth but also balanced scaling of architectural dimensions. EfficientNet, introduced by Tan and Le (2019), proposed a compound scaling strategy that jointly adjusts network depth, width, and input resolution in a principled manner. The original EfficientNet series (B0–B7) demonstrated that this balanced scaling significantly improves accuracy–efficiency trade-offs compared to conventional scaling approaches.

Building upon this framework, EfficientNetV2 (Tan & Le, 2021) introduced architectural refinements such as Fused-MBConv blocks and training-aware optimization strategies to accelerate convergence while maintaining strong performance. The EfficientNetV2 family includes multiple variants, typically categorized as Small (S), Medium (M), and Large (L), each designed to address different computational budgets and deployment scenarios. Smaller variants are suitable for resource-constrained environments, whereas larger variants provide increased representational capacity.

Within this study, EfficientNetV2L was selected as the large-scale representative of the family. Its higher model capacity enables more expressive feature extraction compared to lightweight architectures, while still benefiting from the optimized scaling strategy inherent to the EfficientNet design. Including EfficientNetV2L allows assessment of a modern, compound-scaled architecture alongside classical and lightweight models in the maritime object detection context.

## 3.3. Vision Transformer (ViT)

While most transfer learning approaches in computer vision rely on Convolutional Neural Network architectures, a new paradigm emerged when the Transformer architecture, originally developed for Natural Language Processing at Google Research, was adapted to image analysis.

The Vision Transformer, introduced in 2020, demonstrated that images could be processed as sequences of patches and analysed using self-attention mechanisms rather than convolutional operations. This shift opened the door to attention-based models in visual recognition and object detection tasks.

### 3.3.1. Theoretical Background of ViT

The Vision Transformer (ViT), introduced by Dosovitskiy et al. (2020), extends the original Transformer architecture to visual recognition tasks. The Transformer model was initially proposed by Vaswani et al. (2017) at Google Research for Natural Language Processing (NLP) applications. Its core innovation, the self-attention mechanism[2], enables the model to capture contextual relationships between elements in a sequence without relying on recurrence or convolution. This mechanism significantly improved efficiency and scalability in text analysis and later inspired similar adaptations in computer vision.

Traditional convolutional neural networks (CNNs) operate through local receptive fields and hierarchical feature extraction using convolution and pooling layers. In contrast, ViT divides an image into fixed-size patches and treats each patch as a token, analogous to word embeddings in NLP. Through this approach, ViT models global dependencies among image regions using self-attention rather than local convolutional operations (Dosovitskiy et al., 2020).

In ViT, an input image $I \in R^{H \times W \times C}$ is divided into $N = \frac{HW}{P^2}$ non-overlapping patches of size $P \times P$. Each patch is flattened and linearly projected into an embedding vector, and positional encodings are added to preserve spatial information. The resulting sequence of embeddings is then processed through multiple Transformer encoder layers, each consisting of multi-head self-attention and feed-forward sublayers with residual connections and layer normalisation. The output of the classification token [CLS], provides a global summary of the image and is used for classification or object detection tasks.

Unlike CNNs, which emphasise local spatial relationships, ViT leverages global self-attention to model long-range dependencies across the entire image. This capability allows ViT to capture relationships between distant regions, such as vessels, buoys, and sea clutter, which is particularly valuable in maritime security applications where object appearance, scale, and illumination conditions vary considerably. ViT has demonstrated state-of-the-art performance in large-scale visual recognition tasks when pre-trained on extensive datasets and fine-tuned for downstream applications (Dosovitskiy et al., 2021).

Recent studies have shown the adaptability of Transformer-based architecture for object detection. Li et al. (2023) conducted a comprehensive review and benchmark analysis, highlighting how Transformer models can effectively replace convolutional backbones in detection frameworks and improve robustness under complex and dynamic visual conditions. The global self-attention mechanism enables ViT to capture long-range contextual interactions, enhancing its ability to detect maritime objects under challenging environmental conditions. Despite these advantages, ViT typically requires large-scale datasets or efficient pre-training strategies to achieve optimal performance, and its quadratic computational complexity in self-attention remains a limitation for high-resolution image analysis.

### 3.3.2. Model Architecture and Workflow Diagram

The Vision Transformer (ViT) architecture adapts the Transformer encoder framework to process visual data by converting images into sequences of patches. The model consists of four main components: image tokenization, positional encoding, Transformer encoder layers, and a classification or detection head. Each component plays a specific role in transforming raw pixel information into high-level semantic representations suitable for object detection.

---

[2] A comprehensive explanation of the self-attention mechanism is provided in Hattatoglu and Gocer (2025, pp. 224–225).

In the first stage, the input image $I \in R^{H \times W \times C}$ is divided into $N = \frac{HW}{P^2}$ non-overlapping patches of size $P \times P$. Here H, W and C is Height, Width, (Colour) Channels, respectively. Each patch is flattened into a one-dimensional vector and projected linearly into an embedding of dimension $D$ using a learnable matrix $E \in R^{(P^2 \cdot C) \times D}$. The resulting set of embeddings forms a sequence of patch tokens expressed as $X_p = [x_1 E; x_2 E; \dots; x_N E]$, where $x_i$ represents the $i^{th}$ image patch and $E$ is the embedding matrix.

Because the Transformer is invariant to the input order, positional embeddings $E_{pos} \in R^{N \times D}$ are added to preserve spatial relationships among the patches, resulting in the initial input sequence $Z_0 = X_p + E_{pos}$. This sequence is then passed through a stack of $L$ identical encoder layers, each consisting of two primary submodules: a multi-head self-attention (MSA) mechanism and a feed-forward network (FFN). The self-attention operation enables each token to attend to every other token, thereby capturing global contextual dependencies across the image. It is formally defined as

$$Attention(Q, K, V) = softmax\left(\frac{QK^T}{\sqrt{d_k}}\right)V$$

where $Q, K, V$ denote the query, key, and value matrices, and $d_k$ represents the dimension of the key vector.

The output of the attention module is then passed through a two-layer feed-forward network with a non-linear activation function, typically GELU. To ensure training stability, residual connections and layer normalisation are applied after each sublayer as

$$Z_l' = MSA\big(ln(Z_{l-1})\big) + Z_{l-1}, \;\; Z_l = FFN\big(ln(Z_l')\big) + Z_l'$$

where $ln$ denotes layer normalisation and $l$ indexes the encoder layer index.

Finally, the output representation of the learnable classification token [CLS], or alternatively all patch tokens in detection tasks, is passed to a fully connected head that predicts class probabilities and bounding box coordinates. In object detection frameworks such as DETR or hybrid ViT–CNN models, the ViT architecture outputs multi-scale features that facilitate precise localisation and classification.

The overall workflow of the ViT-based object detection process is illustrated in **Figure 1**, which depicts the step-by-step transformation from input image patches to token embeddings, attention-based feature learning, and final object prediction. This pipeline enables ViT to integrate both local and global contextual cues, which is particularly valuable for detecting maritime objects in complex sea environments.

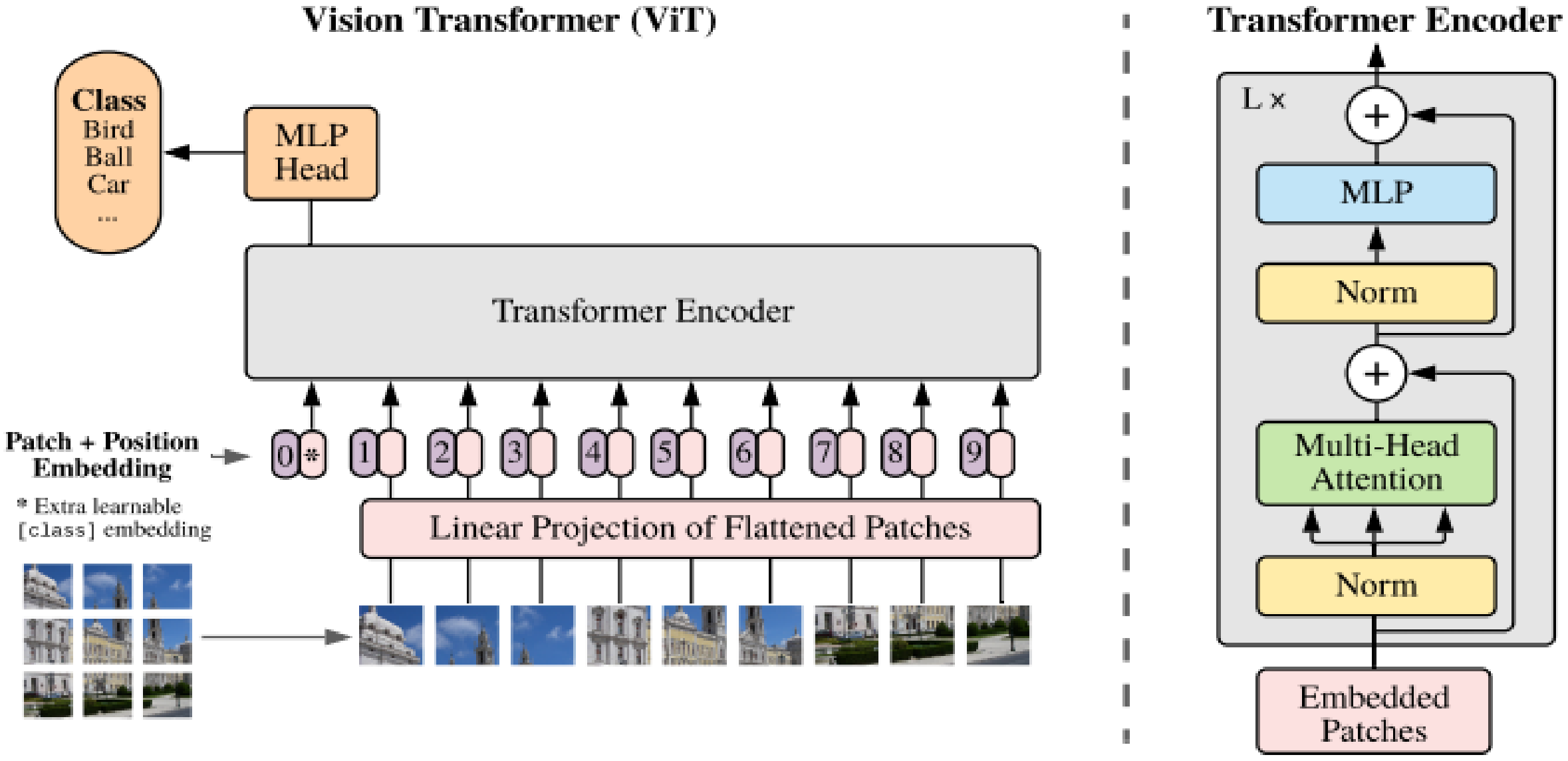

*Figure 1 :* ViT Architecture
*Source :* Dosovitskiy et al. (2021, p. 3).

As seen in Figure 1, the Vision Transformer (ViT) divides an image into equal-sized patches, applies linear embedding to each, and adds positional embedding. The resulting sequence of vectors is then passed to a standard Transformer encoder. For classification tasks, ViT includes a learnable "class token" in the sequence, following the Transformer-based architecture introduced by Vaswani et al. (2017).

## 4. Experimental Design and Comparative Evaluation

In this section, object detection via different computer vision algorithms were conducted.

### 4.1. Dataset Description

The experiments were extended to include a big dataset of 6,468 images from Gocer (2026) Ship Detection Image Dataset. The images were labelled with two classes: Object (1) and Background (0).

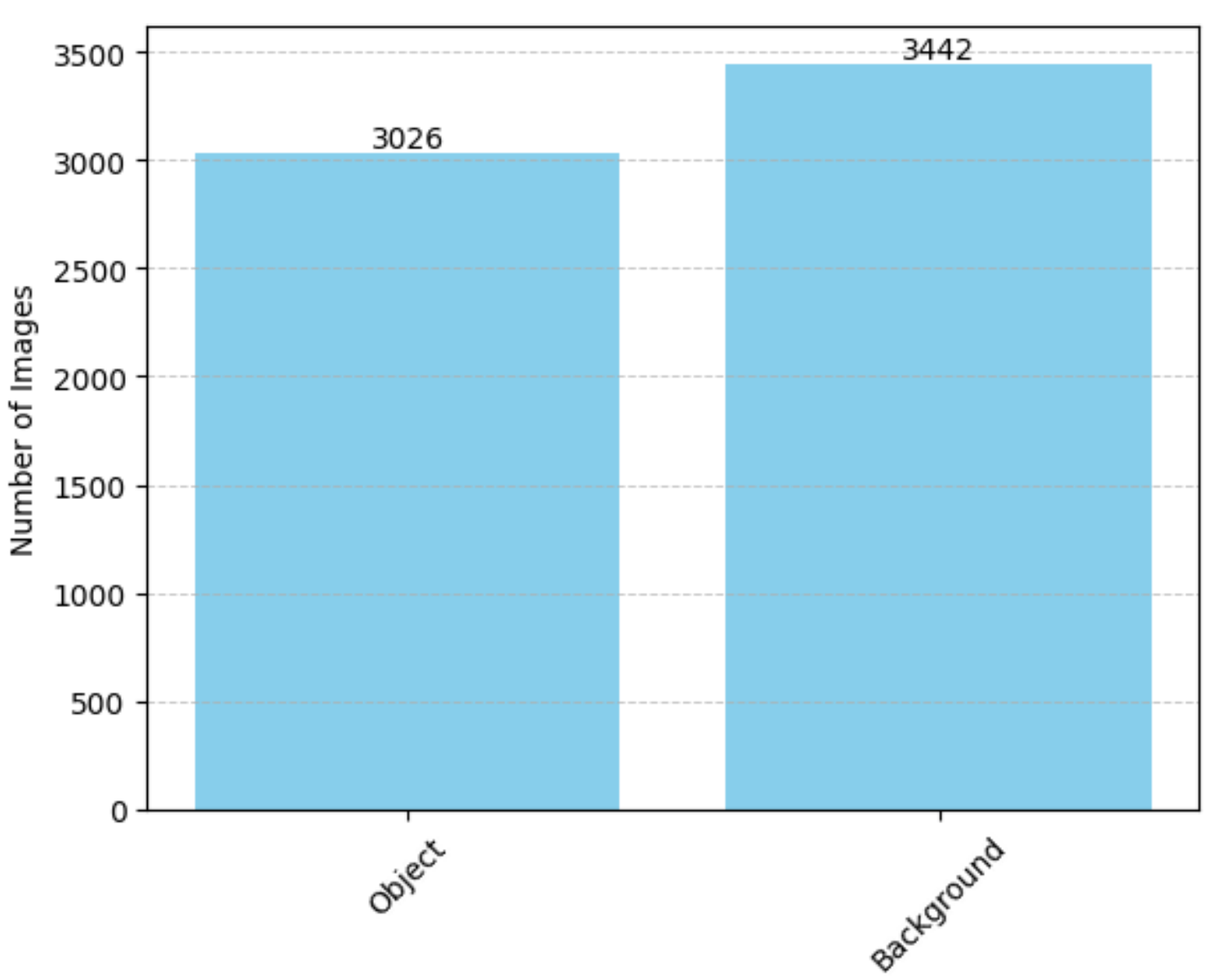

*Figure 2:* Total Number of Images per Class

As it seen at Figure 2, there are total 6468 images in dataset. This dataset was split into 70% for training, 15% for validation and 15% for testing. Then data augmentation processes applied on training data to increase data diversity.

### 4.2. Model Architectures, and Data Preprocessing Strategies

#### 4.2.1 Base CNN Model

Several variations of manually designed CNN architectures were tested and adopted the following configuration in this study as our baseline model for binary image classification. The network accepts input images of size 224 × 224 × 3 and consists of six convolutional blocks with 3 × 3 kernels and ReLU activation. The number of filters progressively increases from 32 to 256 in deeper layers to capture hierarchical features. Batch Normalization was applied in the initial blocks to stabilise training, while MaxPooling layers were used after each block to reduce spatial dimensionality. Dropout with a rate of 0.3 was introduced in deeper layers to mitigate overfitting. The extracted feature maps were flattened and passed to a fully connected layer with 64 neurons, followed by a sigmoid output neuron for binary classification. The model was trained using the Adam optimizer with binary cross-entropy loss, and performance was evaluated using accuracy.

In this study, data augmentation was applied exclusively to the training set to enhance generalisation capability and included random rotation, width and height shifts, shear transformation, zoom, and horizontal flipping. For the validation and test sets, no augmentation was performed; only model-specific preprocessing was applied to preserve the integrity of performance assessment. In addition, data shuffling was disabled[3] during validation and testing to maintain consistent prediction–label alignment and to ensure reproducible, reliable, and unbiased computation of performance metrics.

#### 4.2.2 Transfer Learning Models

In this study, four widely used convolutional neural network (CNN)-based architectures, namely Xception, VGG16, MobileNetV2, and EfficientNetV2L, were employed as transfer learning models. All models were initialized with ImageNet-pretrained weights and utilized as feature extractors. During fine-tuning, the convolutional backbones were frozen to preserve the learned hierarchical representations, while only the newly added fully connected classification layers were trained on the maritime dataset.

To ensure compatibility with the original architectural designs, input images were resized according to each model's default resolution. Specifically, Xception requires an input size of 299 × 299, VGG16 and MobileNetV2 operate on 224 × 224 images, and EfficientNetV2L uses a higher resolution of 480 × 480.

Preprocessing strategies were applied in accordance with the ImageNet pre-training configurations of each model. For Xception and MobileNetV2, pixel values were scaled to the range [-1, 1] using their respective preprocess_input functions. In contrast, VGG16 preprocessing involves channel-wise mean subtraction and conversion from RGB to BGR format. These preprocessing steps ensure that the input data distribution remains consistent with the original large-scale training conditions.

A key distinction arises in the case of EfficientNetV2L. Unlike the other models, EfficientNetV2L includes an internal preprocessing layer (`include_preprocessing=True`), which performs input normalization within the model architecture. This eliminates the need for an external preprocessing function and ensures seamless compatibility with ImageNet pre-training. This feature is specific to

[3] *shuffle = False*

EfficientNetV2 variants, whereas earlier EfficientNet (B0–B7) models require explicit external preprocessing (Hattatoglu and Gocer, 2025, p. 198–238).

Data augmentation was applied exclusively to the training set across all models to improve generalization performance. The augmentation pipeline included random rotation, width and height shifts, shear transformation, zoom, and horizontal flipping. For validation and test sets, no augmentation was performed. Instead, only model-specific preprocessing was applied, and data shuffling was disabled to ensure consistent, reproducible, and unbiased evaluation.

A summary of the preprocessing strategies employed across all transfer learning models is presented in Table 1.

**Table 1.** Summary of preprocessing strategies across the transfer learning models

| Model | Internal Preprocessing | External Preprocesing | Preprocessing Function | Input Size | Normalization Strategy |
|---|---|---|---|---|---|
| Xception | ✗ | ✓ | keras.applications .xception,preprocess | 299 × 299 | Scale to [-1, 1] |
| VGG16 | ✗ | ✓ | keras.applications. .vgg16,preproces_ıгps | 224 × 224 | Mean subtraction, RGB → BGR |
| MobileNetV2 | ✗ | ✓ | keras.applications .mobilenes V2,preproes; | 224 × 224 | Scale to [-1, 1] |
| EfficientNetV2L | ✓ | ✗ | Not required | 480 × 480 | Internal scaling to [-1, 1] |

**Notes:** ✓: indicates applied, X: indicates not applied. EfficientNetV2L incorporates a built-in preprocessing layer (`include_preprocessing=True`), whereas other models require explicit external preprocessing to maintain consistency with ImageNet pre-training.

Table 1 provides a comparative overview of the preprocessing configurations employed across the transfer learning models. It emphasizes model-specific design choices, including input size requirements, normalization schemes, and the distinction between internal and external preprocessing strategies to maintain compatibility with ImageNet pre-training. The overall workflow of the transfer learning pipeline is illustrated in Figure 3.

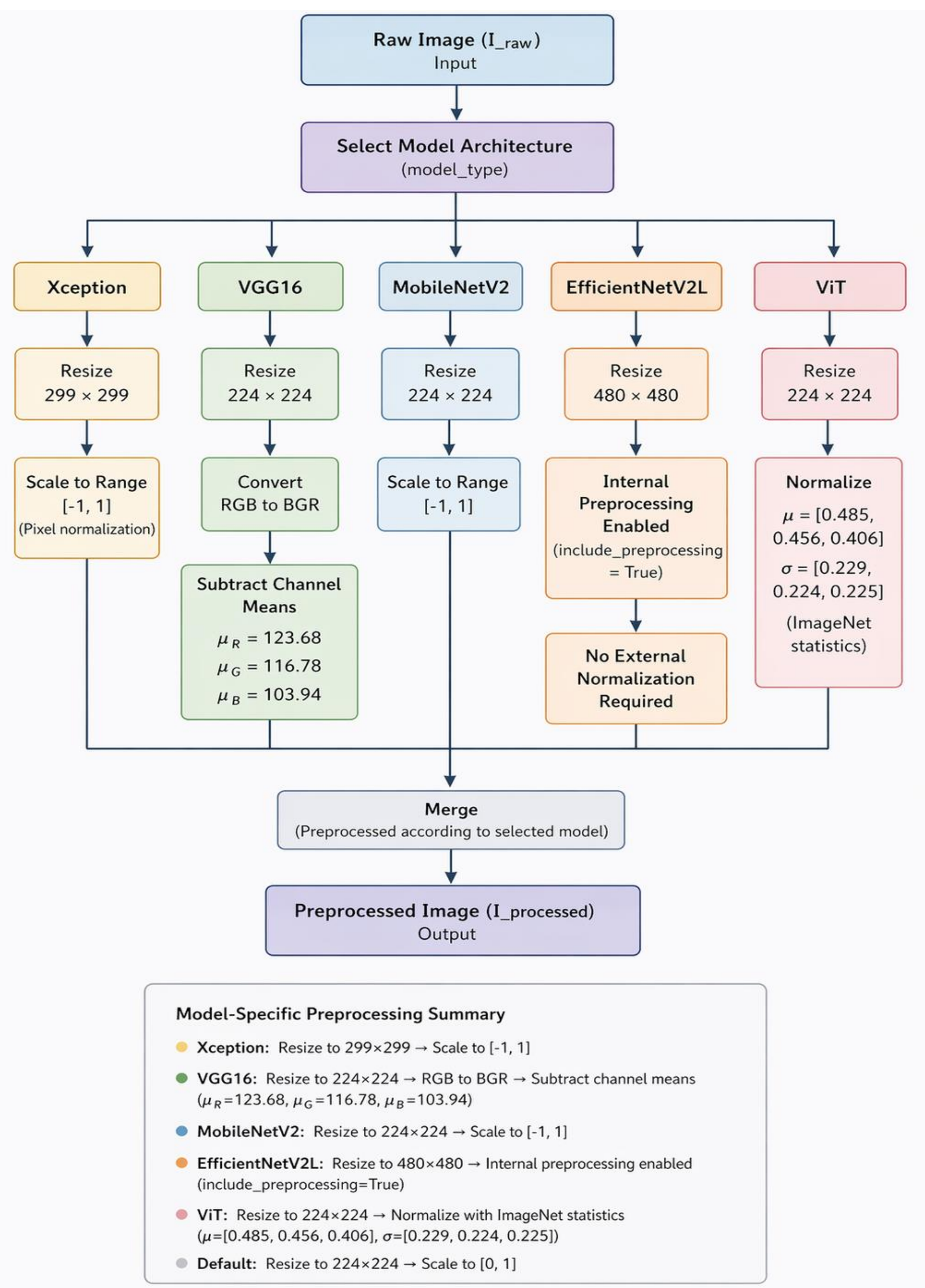

*Figure 3:* Transfer learning workflow illustrating preprocessing strategies

As illustrated in Figure 3, the overall workflow follows a unified transfer learning pipeline in which input images are resized and processed through either external preprocessing strategies (*Xception*, *VGG16*, *MobileNetV2*) or an internal preprocessing mechanism (*EfficientNetV2L*). The pipeline further incorporates data augmentation exclusively during the training phase, while maintaining frozen feature extraction backbones and optimizing only the classification layers, thereby ensuring both architectural consistency with ImageNet pre-training and efficient model adaptation to the target dataset.

**4.2.3 ViT**

In this study, we employ the ImageNet-pretrained ViT-Base (Patch16, 224×224) model (google/vit-base-patch16-224) and fine-tune its weights on the maritime dataset. All input images were processed according to the model's architectural requirements. The pretrained ViT expects fixed-resolution RGB images of size 224×224 pixels and normalization aligned with ImageNet statistics. Accordingly, images were resized and normalized using the mean and standard deviation provided by the official model configuration.

For the training set, data augmentation was applied to improve generalization and mitigate overfitting. Specifically, images were randomly resized and cropped to 224×224 pixels using a scale range of 0.8 to 1.0, followed by random horizontal flipping with probability 0.5. To enhance robustness against variations in illumination and colour conditions, mild photometric transformations were introduced via colour jitter, including controlled adjustments to brightness, contrast, saturation, and hue. After augmentation, images were converted to tensors and normalized using ImageNet mean and standard deviation values.

For validation and test sets, no stochastic augmentation was applied to ensure deterministic evaluation. Images were resized to 256 pixels on the shorter side, centre-cropped to 224×224 pixels, and normalized using the same ImageNet statistics as the training set. This consistent preprocessing strategy ensures compatibility with the pretrained backbone while preventing information leakage during model evaluation.

The dataset was organized in a class-wise directory structure and loaded using the PyTorch ImageFolder utility, ensuring consistent label indexing across training, validation, and test splits. The proposed vision-based ship detection architecture is designed as an efficient end-to-end processing pipeline that transforms raw UAV/USV camera recordings into actionable classification outputs. As illustrated in Figure 4, the system follows a streamlined "*Ingest → Detect → Output*" paradigm, optimized for robust performance and generalization under diverse maritime conditions.

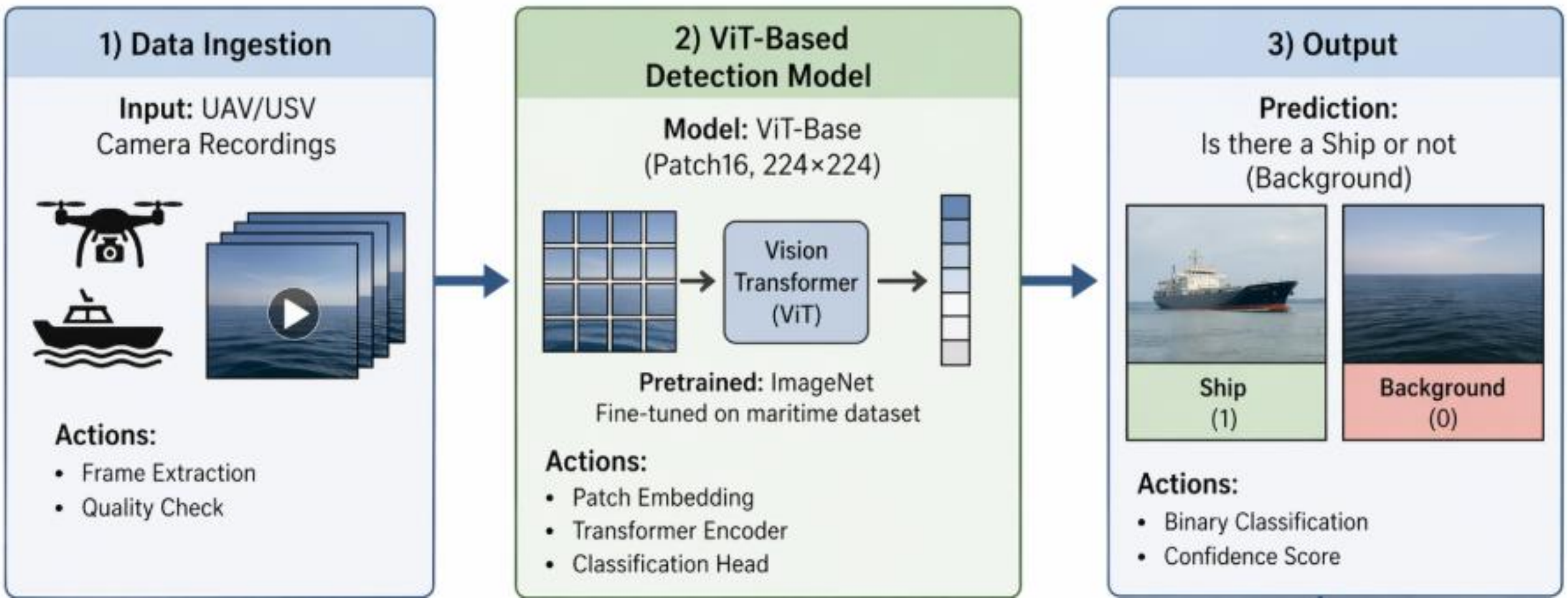

*Figure 4:* ViT Functional Intelligence Architecture

The pipeline in Figure 4 consists of three primary functional stages:

1. **Data Ingestion (Preprocessing):** Raw visual data collected from UAV and USV platforms is ingested and converted into frame-level inputs. Key preprocessing steps include frame extraction, quality control, resizing, and normalization according to ImageNet standards. This stage ensures consistency in input representation and prepares the data for compatibility with the pretrained Vision Transformer architecture.
2. **ViT-Based Detection Model:** The processed image frames are fed into a fine-tuned Vision Transformer (ViT-Base, Patch16, 224×224) model. Leveraging global self-attention mechanisms, the ViT model captures both local and long-range dependencies within maritime scenes, enabling robust discrimination between ships and complex sea backgrounds. The model is pretrained on ImageNet and subsequently fine-tuned on the maritime dataset to enhance domain-specific feature learning. The output of this stage is a probabilistic classification score indicating the presence of a ship.
3. **Output (Classification):** In the final stage, the model produces a binary decision for each frame, classifying it as either **Ship (1)** or **Background (0)**, along with an associated confidence score. These predictions can be directly utilized for downstream monitoring, surveillance, or decision-support systems.

This streamlined architecture operates solely on visual information, enabling efficient and scalable deployment. By focusing on high-level feature representation through transformer-based learning, the system demonstrates strong generalization capability across varying weather conditions, illumination levels, and sea states, making it suitable for real-world maritime surveillance applications.

### 4.3 Comparative Performance Analysis and Model Selection

In this research, the following methods were implemented to compare their performance and choose the best model for object detection: (i) Base (manual) CNN model, (ii) transfer learning algorithms such as Xception, VGG16, MobileNetV2, EfficientNetV2L and Vision Transformer (ViT). All models were trained for up to 500 epochs with early stopping (patience = 10). After training the models, their video processing performance was tested on a 37 second video. Table 2 presents a comprehensive comparison of the models developed in this experiment.

**Table 2.** Comprehensive comparison of models developed in extended experiments.

| Model | Number of Parameters | Training Accuracy | Testing Accuracy | Type I Error | Type II Error | Model Size | Computing Time for 37 Sec. Video |
|---|---|---|---|---|---|---|---|
| **Base CNN** | 1,974,849 | 0.99 | 0.99 | 6 | 1 | 23.2 MB | 48.08 Sec. |
| **XCeption** | 20,992,681 | 0.99 | 1 | 3 | 2 | 83.3 MB | 31.85 Sec |
| **VGG16** | 14,747,585 | 0.99 | 0.99 | **2** | **1** | 57.6 MB | 32.85 Sec |
| **MobileNetV2** | 2,340,033 | 0.99 | 0.99 | 4 | 3 | 10 MB | 34.08 Sec |
| **EfficientNetV2L** | 117,828,897 | 0.99 | 1 | 2 | 4 | 462.5 MB | 183.23 Sec |
| **Vision Transformer (ViT)** | 85,800,194 | **1** | **1** | **1** | **3** | 335.2 MB | **30.97 Sec** |

Table 2 shows training and testing accuracies, the number of Type I and Type II errors, the size of the model in megabytes, and time the models took to process a 37-second recorded video having multiple scenarios of ship(s) and plain water. Accordingly, ViT model achieved the highest accuracies (100%) not only on the training and testing sets but also on the recorded video with minimum processing time (30.97 Seconds) with 100% success rate. However, the size of ViT model (335.2 MB) is larger than other models, except for EfficientNetV2L (462.5 MB) which remained the largest model with the highest processing time (183.23 Seconds) on the recorded video. At the end of these analyses, we decided ViT is the best model for this task. An early stopping mechanism with a patience of 5 was incorporated into the architecture, and the model was trained for up to 500 epochs.

Figure 5 and 6 illustrate the training convergence graph and validation confusion matrix ViT model. It is obvious that ViT model converged with variance in performances on training and validation sets, except for ViT model which achieved similar high-level performance (100% accuracy) on both the sets (seen (train) and unseen (test) data). Moreover, from the confusion matrix, it is clear that the ViT model made very few false positive and false negative errors on validation set.

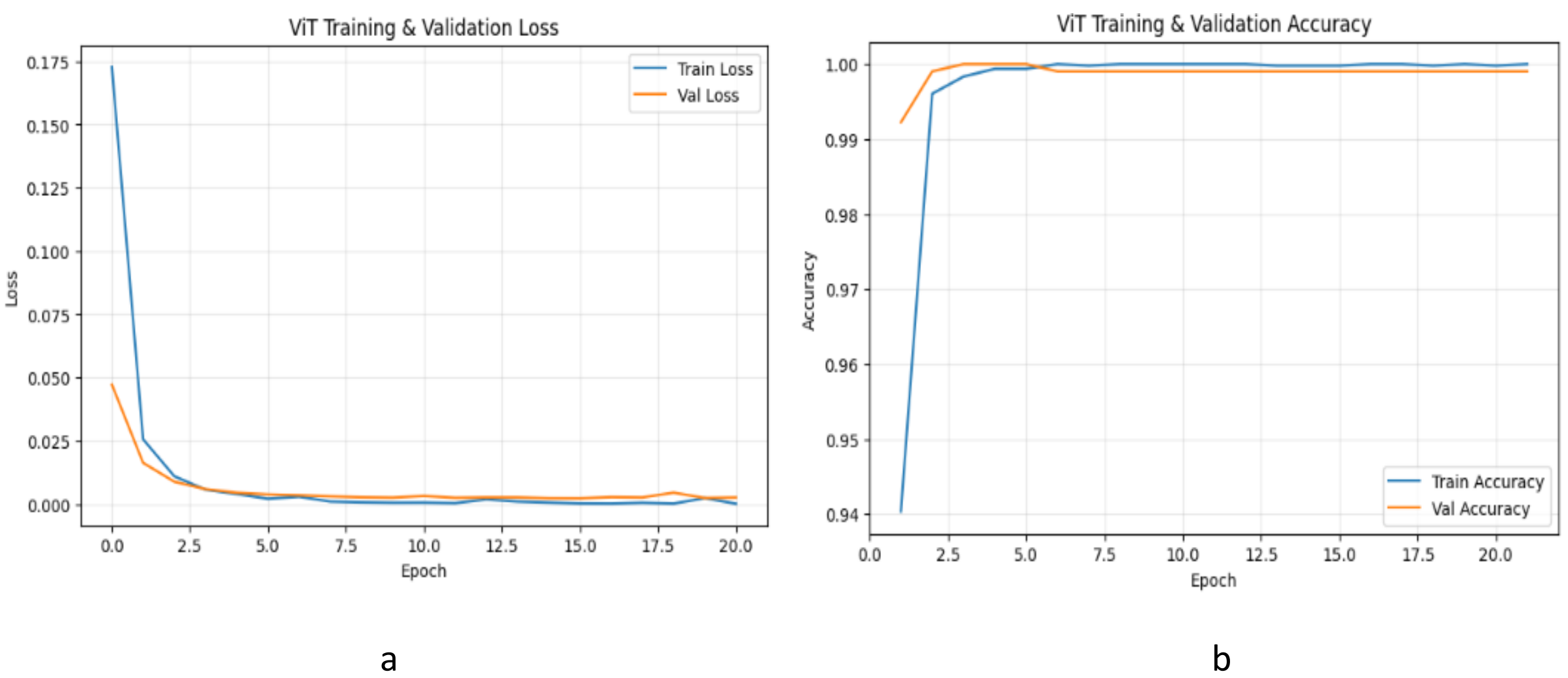

*Figure 5: Loss and Accuracy Graphs of ViT model*

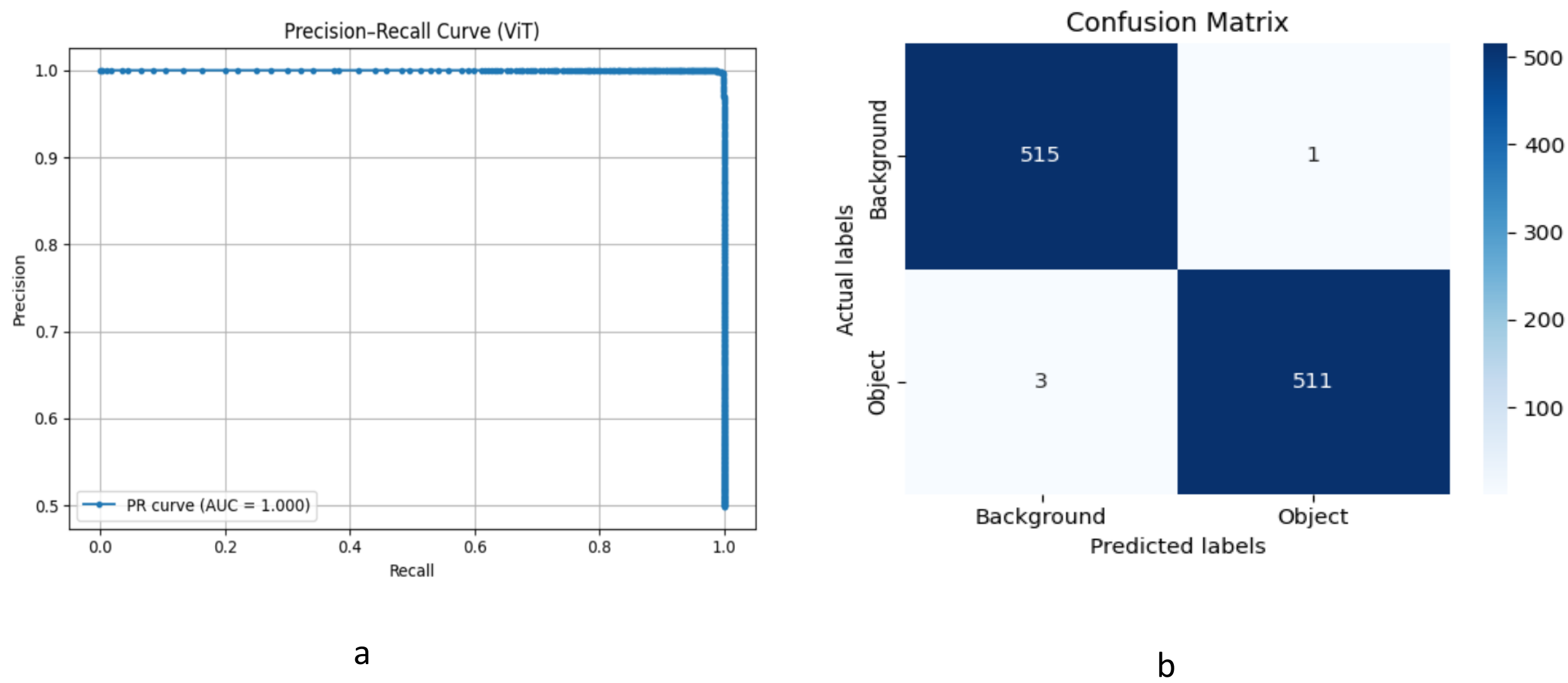

*Figure 6: Precision - Recall Graphs and Confusion Matrix of ViT model*

As we can see in Figure 5 and 6, there are around zero error (loss) at Figure 5a. Figure 6a shows AUC = 1, meaning the ViT model has 100% successful. Lastly Figure 6b proves there are one Type I error and three Type II error. Shortly, ViT model has high performance for object detection.

### 4.4 Inferences from ViT

The results obtained from testing the trained ViT model on unseen background and ship images are presented in Figures 7 and 8.

**Figure 7. Inferences from ViT on Background Images**

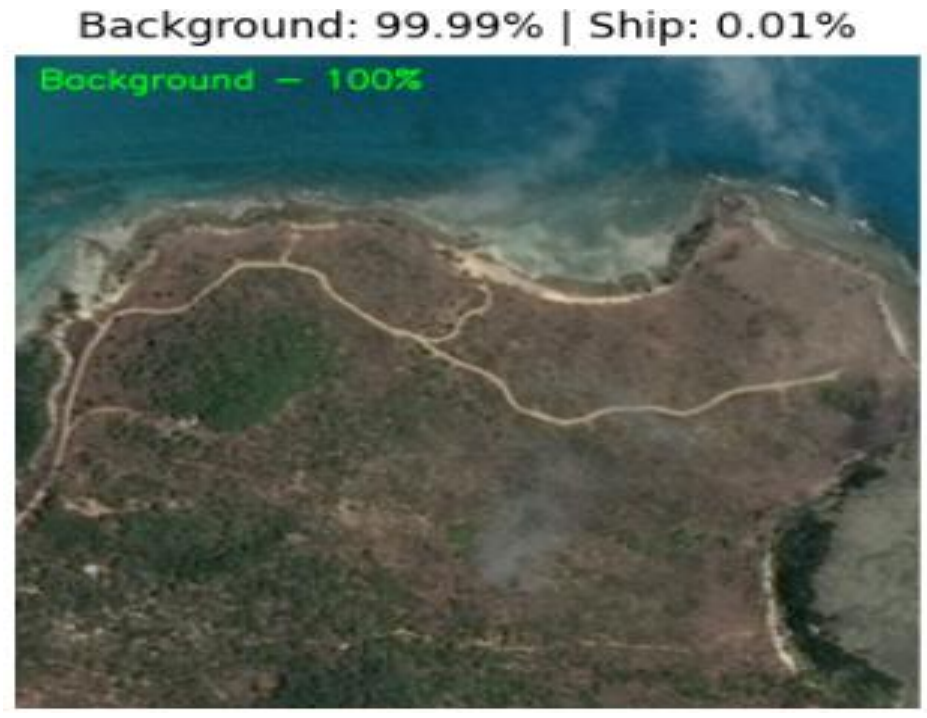

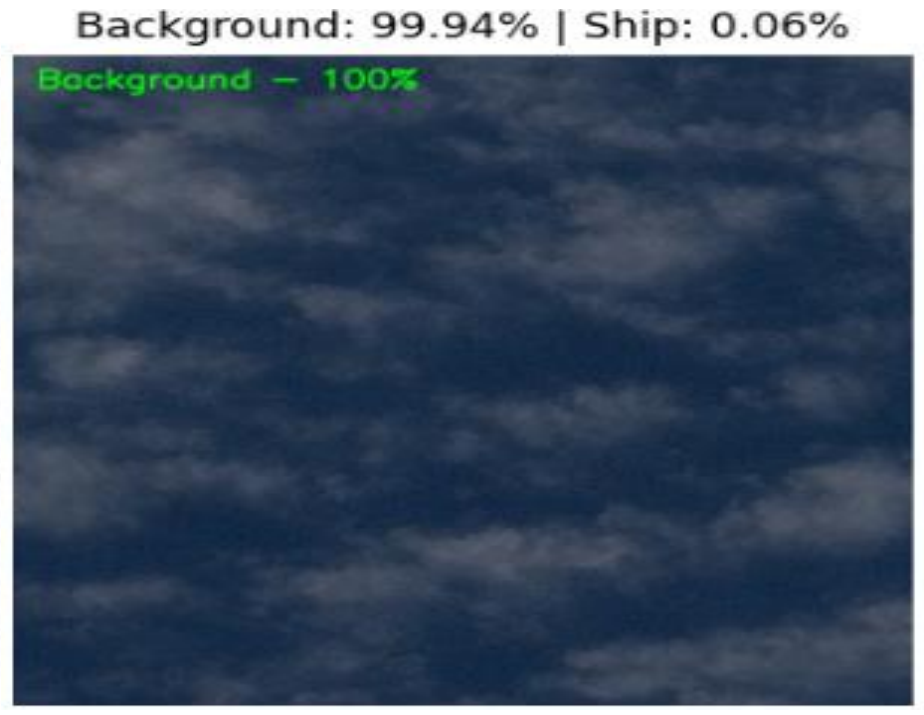

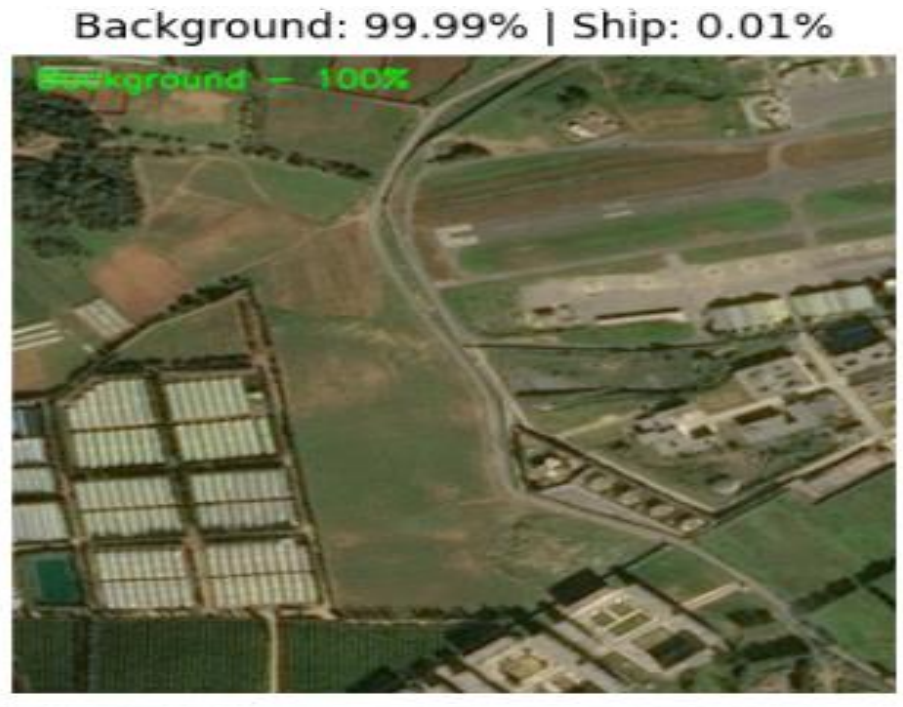

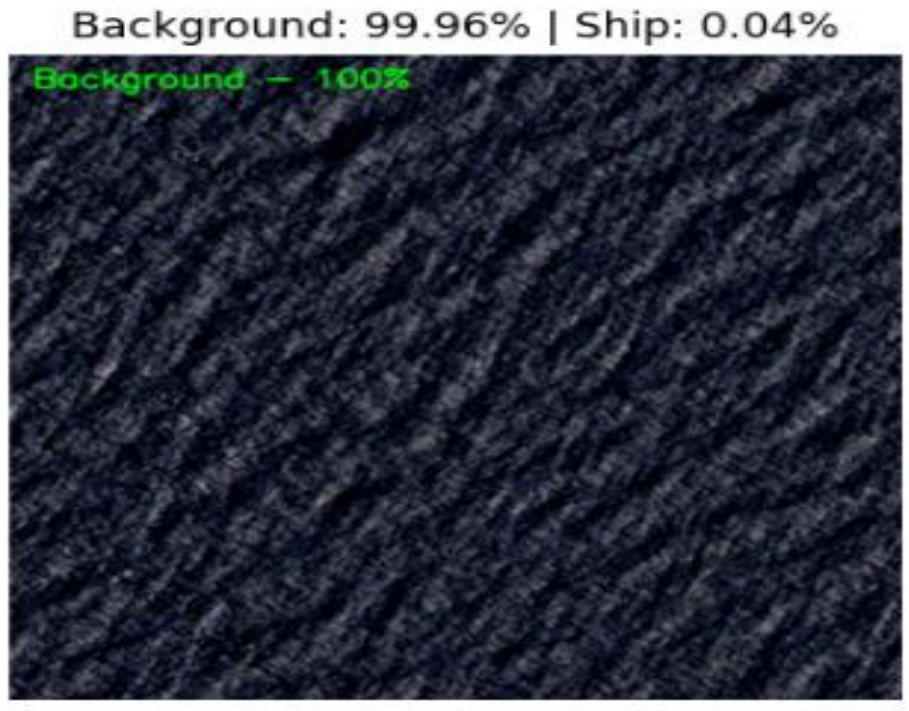

**Figure 8. Inferences from ViT on Object (Ship) Images**

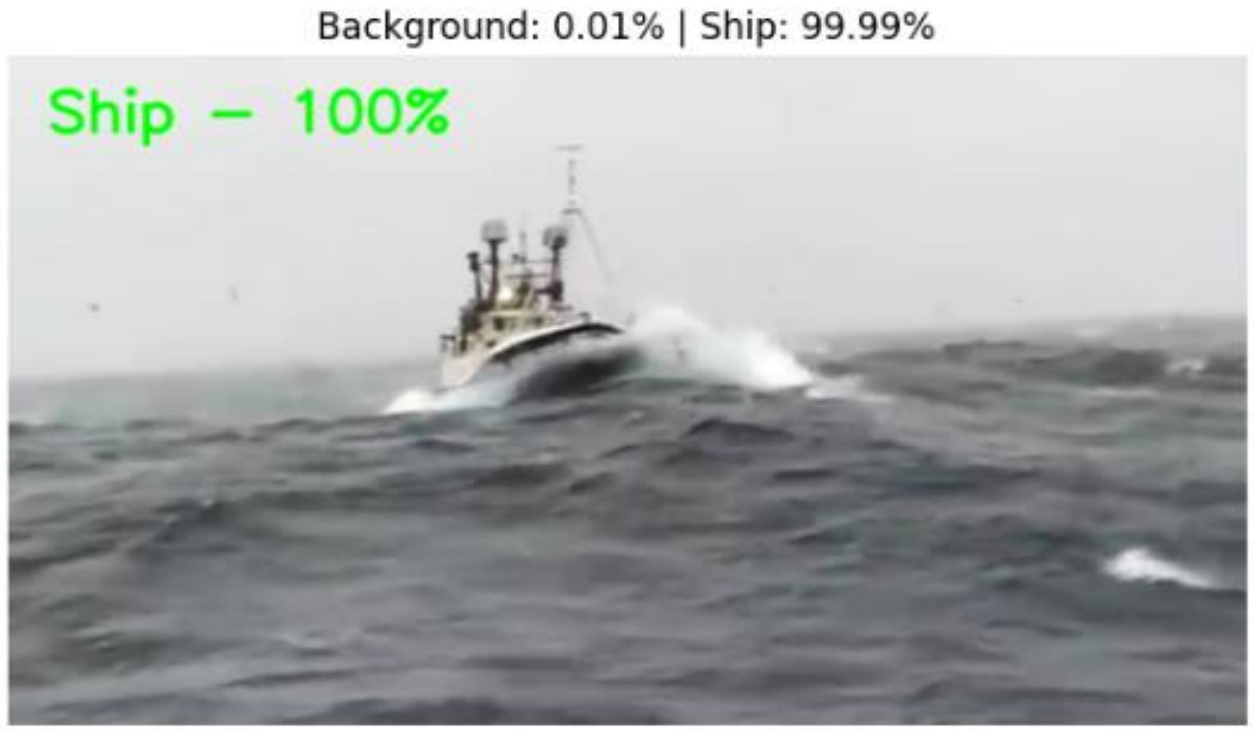

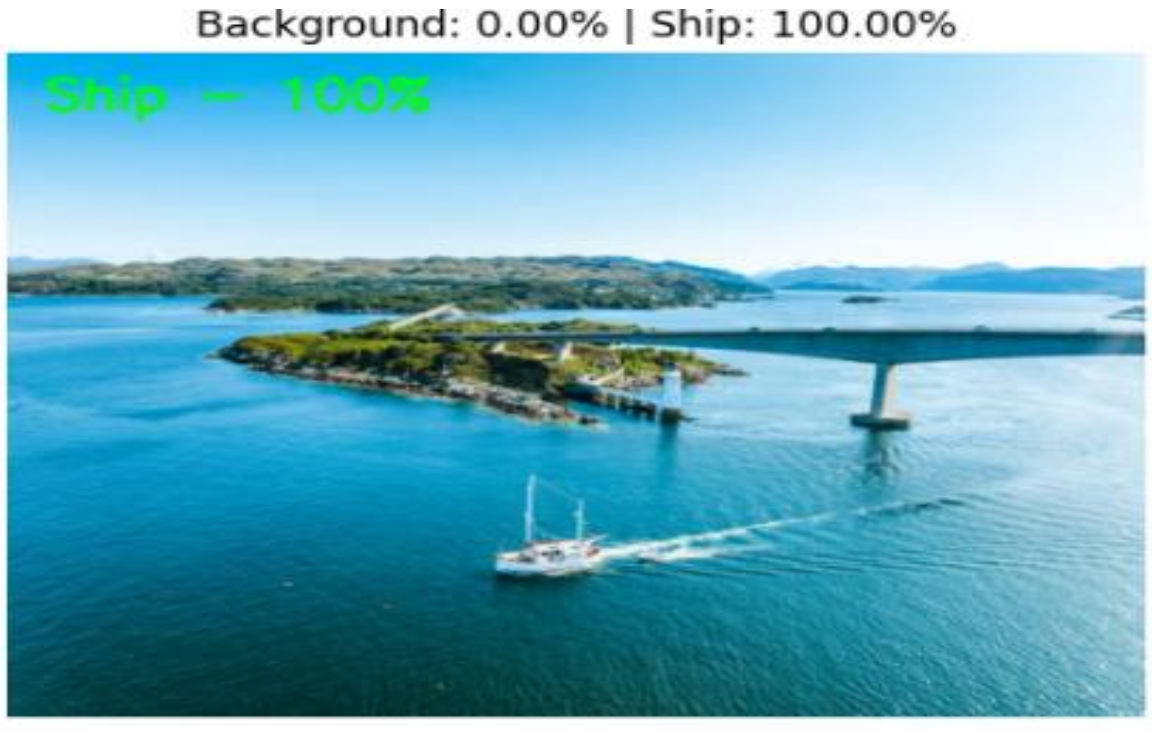

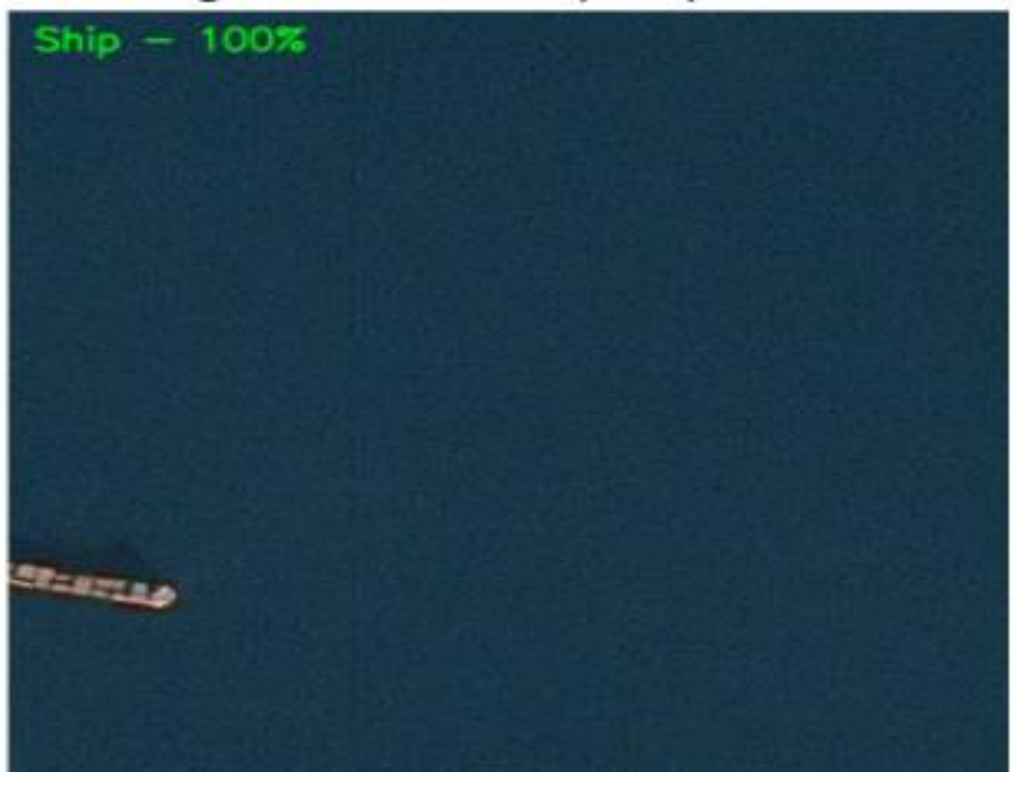

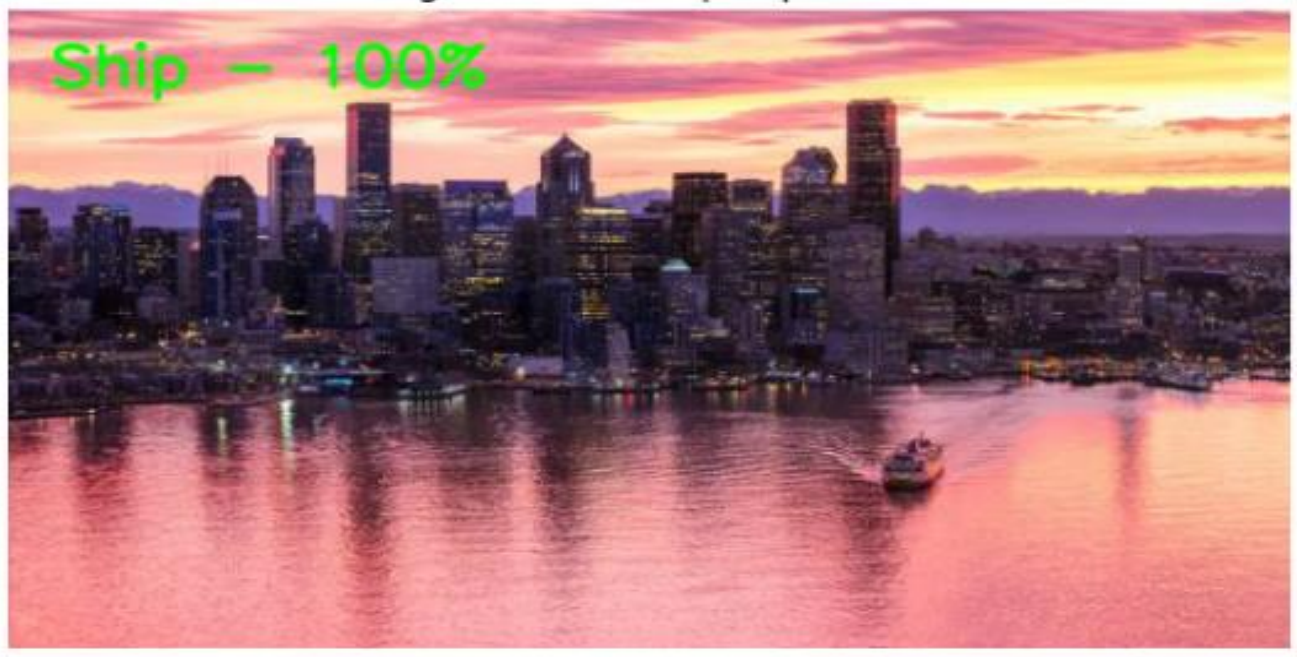

Results in Figures 7 and 8 indicate that the trained ViT model achieves strong performance across evaluation metrics, demonstrating robust accuracy and generalization. The model was trained and tested on images captured under diverse weather conditions and varying sunlight effects, which further enhances its generalization capability.

## 5. Discussion

The experimental results obtained in this study reveal several important insights regarding the performance of different computer vision architectures for maritime object detection.

First, the results indicate that model performance is strongly influenced by the architectural design of the learning framework. Lightweight convolutional neural network architectures such as MobileNetV2 demonstrate clear advantages in terms of model size and deployability on resource-constrained devices. These properties make such models attractive for edge-based monitoring systems where computational capacity and memory resources are limited. However, the findings also suggest that lightweight models may exhibit relatively higher error rates when processing complex maritime scenes. This limitation may be attributed to the reduced representational capacity of compact architectures, which can make it more challenging to capture subtle visual patterns in ocean environments characterized by waves, reflections, and varying lighting conditions.

Second, the results highlight the importance of balancing detection accuracy and computational efficiency in real-world maritime monitoring applications. Models such as VGG16 show relatively strong detection performance while maintaining moderate computational requirements. This balance suggests that medium-complexity architectures can serve as a practical compromise in systems where both reliability and computational efficiency are important operational constraints.

A particularly notable finding of this study is the superior performance of the Vision Transformer (ViT) model. The transformer-based architecture achieved the highest accuracy levels while simultaneously minimizing both Type I and Type II errors. In addition, the ViT model demonstrated the fastest processing time when applied to recorded video data. These results support recent findings in the computer vision literature suggesting that transformer-based architectures can capture global contextual relationships in images more effectively than traditional convolutional networks. In maritime environments, where objects may appear at different scales and positions within complex backgrounds, this global attention mechanism may provide a significant advantage.

Another important aspect highlighted in this study is the critical role of data preprocessing in transfer learning-based computer vision systems. In many applied studies, preprocessing pipelines are either insufficiently described or implemented inconsistently across different architectures. Such inconsistencies can lead to misleading performance comparisons and reproducibility challenges. In this work, the preprocessing procedures were systematically adapted for each architecture, ensuring that the transfer learning models were applied under appropriate input configurations. This methodological transparency contributes to more reliable comparisons between architectures and provides practical guidance for future implementations.

From an application perspective, the results suggest that model selection should be guided not only by predictive accuracy but also by deployment conditions and operational constraints. In maritime monitoring systems deployed on drones, autonomous vessels, or edge-based surveillance platforms, lightweight models may be preferable due to their reduced computational requirements. Conversely, in centralized monitoring systems equipped with higher computational capacity, transformer-based architectures such as ViT may provide superior detection reliability and operational performance.

Overall, the findings demonstrate that no single architecture is universally optimal for all deployment environments. Instead, the optimal model choice depends on a combination of accuracy requirements, computational resources, processing speed, and acceptable error levels. By evaluating these factors jointly, this study provides a more comprehensive understanding of how different deep learning architectures perform under realistic maritime monitoring conditions.

## 6. Conclusion

This study investigated the performance of six different computer vision algorithms for object detection on sea surface imagery, including a base CNN architecture, four transfer learning models (Xception, VGG16, MobileNetV2, and EfficientNetV2L), and a Vision Transformer (ViT) model. The models were evaluated using multiple performance indicators, including training and testing accuracy, Type I and Type II errors, model size, and computational processing time on recorded video sequences.

The experimental results demonstrate that the models exhibit different strengths depending on the operational context and computational constraints. Lightweight architectures such as MobileNetV2 provide advantages in terms of model size and potential deployability on mobile or edge devices with limited hardware capacity. However, these models may produce relatively higher error rates and longer processing times when applied to complex maritime scenes. In contrast, architectures such as VGG16 offer a more balanced trade-off between detection accuracy and computational efficiency, making them suitable for systems with moderate computational resources.

Among the evaluated approaches, the Vision Transformer (ViT) model achieved the best overall performance. The model reached 100% accuracy on both training and testing datasets, produced the lowest number of Type I and Type II errors, and demonstrated the fastest processing time on the recorded video data. These results indicate that transformer-based architectures can provide highly reliable performance for maritime object detection tasks when sufficient computational resources are available. Although the ViT model has a relatively larger model size compared with several CNN-based architectures, the experimental findings suggest that the accuracy and reliability benefits outweigh the additional computational requirements in high-capacity systems.

Beyond the empirical findings, this research contributes to literature in two important ways. First, the study provides a clear and structured description of the data preprocessing pipeline required when applying transfer learning in different deep learning architectures. Data preprocessing is a critical step in computer vision applications, yet it is frequently implemented incorrectly or insufficiently reported in transfer learning studies. In this work, the preprocessing procedures are explicitly described and adapted for different model architectures, providing a methodological reference that can improve the reliability and reproducibility of similar applications. Second, the study offers a scenario-oriented comparison of multiple detection models, demonstrating how model selection should consider not only accuracy but also computational efficiency, processing speed, and error costs.

From a practical perspective, the results indicate that the optimal model choice depends strongly on the deployment environment. Lightweight models may be preferable for edge-based maritime monitoring systems, while transformer-based architectures such as ViT provide superior performance for high-capacity surveillance platforms where detection reliability is critical.

## 7. Future Research

Future work may extend this research by evaluating the proposed models on larger and more diverse maritime datasets, investigating real-time deployment scenarios, and exploring hybrid architectures that combine the efficiency of lightweight convolutional networks with the representational capabilities of transformer-based models. Such developments may further improve the robustness and scalability of intelligent maritime monitoring systems.

**Acknowledgement:** During the preparation of this work, the authors used ChatGPT (24 May 2023 version, OpenAI, San Francisco, CA, USA) to assist in language polishing, thereby improving the clarity and readability of the manuscript. After using this tool/service, the authors reviewed and edited the content as needed and took full responsibility for the content of the publication.

**Funding Statement:** This project has received funding from the European Union's Horizon Europe research and innovation programme under the Grant Agreement 101168489.

**Availability of Data and Materials:** In this study Gocer (2026) "Ship Detection Image Dataset" was used to analyses. It is available on doi: 10.17632/4pjzdc55nt.1

**Ethics Approval:** Not applicable.

**Conflicts of Interest:** The authors declare no conflicts of interest.

## References

Bochkovskiy, A., Wang, C. Y., & Liao, H. Y. M. (2020). *YOLOv4: Optimal speed and accuracy of object detection. https://arxiv.org/abs/2004.10934*.

Carion, N., Massa, F., Synnaeve, G., Usunier, N., Kirillov, A. and Zagoruyko, S. (2020). *End-to-End Object Detection with Transformers (DETR). European Conference on Computer Vision (ECCV 2020).* https://arxiv.org/abs/2005.12872

Chollet, F. (2017). Xception: Deep Learning with Depthwise Separable Convolutions. CVPR. https://openaccess.thecvf.com/content_cvpr_2017/papers/Chollet_Xception_Deep_Learning_CVPR_2017_paper.pdf

Dosovitskiy, A., Beyer, L., Kolesnikov, A., Weissenborn, D., Zhai, X., Unterthiner, T., Dehghani, M., Minderer, M., Heigold, G., Gelly, S., Uszkoreit, J. and Houlsby, N. (2021). *An image is worth 16x16 words: Transformers for image recognition at scale.* https://arxiv.org/pdf/2010.11929

Dumitriu, A., Tatui, F., Miron, F., Ionescu, R. T. and Timofte, R. (2025). Rip Current Segmentation: A Novel Benchmark and YOLOv8 Baseline Results. https://arxiv.org/html/2504.02558v1.

Fan, W., Zhou, F., Bai, X., Tao, M. and Tian, T. (2019). Ship Detection Using Deep Convolutional Neural Networks for PolSAR Images. Remote Sensing, 11(23), 2862; https://doi.org/10.3390/rs11232862

Gocer, I. (2026). "Ship Detection Image Dataset", Mendeley Data, V1, doi: 10.17632/4pjzdc55nt.1

Hattatoglu, F. and Gocer, I. (2025). *Easy CNN & COMPUTER VISION For Everyone*. Amazon Book, ISBN-13: 979-8317445508, https://www.amazon.com/dp/B0F4K55M3R

Herfandi, H., Sitanggang, O. S., Nasution, M. R. A., Nguyen, H. and Jang, Y. M. (2024). Real-Time Patient Indoor Health Monitoring and Location Tracking with Optical Camera Communications on the Internet of Medical Things. Applied Sciences, 14(3), 1153; https://doi.org/10.3390/app14031153

Howard, A.G., Zhu, M., Chen, B., Kalenichenko, D., Wang, W., Weyand, T., Andreetto, M. and Adam, H. (2017). MobileNets: Efficient Convolutional Neural Networks for Mobile Vision Applications. *https://aiins.cs.nthu.edu.tw/wp-content/uploads/2018/08/LW_180522.pdf*

Howard, A., Sandler, M., Chu, G., Chen, L-C., Chen, B., Tan, M., Wang, W., Zhu, Y., Pang, R., Vasudevan, V., Le, Q.V., Adam, H. (2019). *Searching for MobileNetV3.* ICCV. https://arxiv.org/abs/1905.02244

Huang, Z. , Li, L. , Krizek, G. and Sun, L. (2023) Research on Traffic Sign Detection Based on Improved YOLOv8. *Journal of Computer and Communications*, 11, 226-232. doi: 10.4236/jcc.2023.117014.

Ioffe, S., & Szegedy, C. (2015). Batch Normalization: Accelerating deep network training by reducing internal covariate shift. *Proceedings of the 32nd International Conference on Machine Learning*, PMLR 37:448-456, 2015. https://proceedings.mlr.press/v37/ioffe15.html

Jia, C., Wang, D., Liu, J. and Deng, W. (2024). Performance Optimization and Application Research of YOLOv8 Model in Object Detection. Academic Journal of Science and Technology, 10(1), **DOI:** https://doi.org/10.54097/p9w3ax47

Jocher, G., Chaurasia, A. and Qiu, J. (2023) YOLO by Ultralytics. https://github.com/ultralytics/ultralytics.

Juan, N. P., Valdecantos, V. N. and Troch, P. (2025). Advancing artificial intelligence in ocean and maritime engineering: trends, progress, and future directions. *Ocean Engineering,* 339(1), 122077. https://doi.org/10.1016/j.oceaneng.2025.122077

Karst, J., McGurrin, R., Gavin, K., Luttrell, J., Rippy, W., Coniglione, R., McKenna, J. and Riedel, R. (2025). Enhancing Maritime Domain Awareness Through AI-Enabled Acoustic Buoys for Real-Time Detection and Tracking of Fast-Moving Vessels. *Sensors (Basel)*, 25(6):1930*. doi: 10.3390/s25061930*.

Khan, S., Naseer, M., Hayat, M., Zamir, S. W., Khan, F. S. and Shah, M. (2022). *Transformers in Vision: A Survey. ACM Computing Surveys*, 54(10), 1–41. https://doi.org/10.1145/3505244

Kristan, M., Leonardis, A., Matas, J., et al. (2016). *The Visual Object Tracking VOT2016 Challenge Results. European Conference on Computer Vision (ECCV) Workshops.* https://ora.ox.ac.uk/objects/uuid:0bcfadda-c6fe-4db9-bb49-b9d21c9881e3/files/s2j62s679p

Krizhevsky, A., Sutskever, I., & Hinton, G. (2012). ImageNet classification with deep convolutional neural networks. *Advances in Neural Information Processing Systems (NeurIPS)*. https://papers.nips.cc/paper_files/paper/2012/file/c399862d3b9d6b76c8436e924a68c45b-Paper.pdf

LeCun, Y., Bottou, L., Bengio, Y., & Haffner, P. (1998). Gradient-based learning applied to document recognition. *Proceedings of the IEEE*, 86(11), 2278–2324. https://ieeexplore.ieee.org/document/726791

Li, Y., Mao, H., Girshick, R. and He, K. (2022). *Exploring Plain Vision Transformer Backbones for Object Detection (ViTDet). European Conference on Computer Vision (ECCV 2022).* https://arxiv.org/abs/2203.16527

Li, Y., Miao, N., Ma, L., Shuang, F. and Huang, X. (2023). *Transformer for object detection: Review and benchmark.* Engineering Applications of Artificial Intelligence, 126(C), 107021. https://doi.org/10.1016/j.engappai.2023.107021

Li, Y., Jia, Z., Zhang, X., Yang, F. and Yang, X. (2025). *A lightweight and efficient ship detection model for complex maritime environments. Applied Ocean Research, 164, 104797*. https://doi.org/10.1016/j.apor.2025.104797

Lin, T. Y., Goyal, P., Girshick, R., He, K., & Dollar, P. (2017). *Focal loss for dense object detection. Proceedings of the IEEE International Conference on Computer Vision (ICCV)*, 2980–2988. https://openaccess.thecvf.com/content_iccv_2017/html/Lin_Focal_Loss_for_ICCV_2017_paper.html

Liu, W., Anguelov, D., Erhan, D., Szegedy, C., Reed, S., Fu, C. Y. and Berg, A. C. (2016). *SSD: Single Shot MultiBox Detector.* In: Leibe, B., Matas, J., Sebe, N., Welling, M. (eds) Computer Vision – ECCV, vol 9905. Springer, Cham. https://doi.org/10.1007/978-3-319-46448-0_2

Liu, L., Ye, J., Li, S., Hu, S. and Wang, Q. (2022). A Novel Machine Learning Algorithm for Cloud Detection Using AERI Measurement Data. *Remote Sensing*, 14(11), 2589. https://doi.org/10.3390/rs14112589

Ma, M., Chen, J., Liu, W. and Yang, W. (2018). Ship Classification and Detection Based on CNN Using GF-3 SAR Images. *Remote Sensing*, 10(12), 2043. https://doi.org/10.3390/rs10122043

Munim, Z. H. (2020). Big data and artificial intelligence in the maritime industry: a bibliometric review and future research directions. *Maritime Policy & Management*, 47(5), 609-624. https://doi.org/10.1080/03088839.2020.1788731

Prasad, D. K., Rajan, D., Rachmawati, L., Rajabaly, E. and Quek, C. (2017). *Video Processing from Electro-Optical Sensors for Object Detection and Tracking in a Maritime Environment: A Survey. IEEE Transactions on Intelligent Transportation Systems*, 18(8), 1993–2016. https://ieeexplore.ieee.org/document/7812788

Redmon, J., Divvala, S., Girshick, R. and Farhadi, A. (2016). *You Only Look Once: Unified, Real-Time Object Detection. IEEE Conference on Computer Vision and Pattern Recognition (CVPR 2016).* https://doi.org/10.1109/CVPR.2016.91

Ren, S., He, K., Girshick, R., & Sun, J. (2015). *Faster R-CNN: Towards Real-Time Object Detection with Region Proposal Networks. Advances in Neural Information Processing Systems (NeurIPS 2015).* https://arxiv.org/abs/1506.01497

Ribeiro, M., Damas, B. and Bernardino, A. (2022). *Real-Time Ship Segmentation in Maritime Surveillance. Sensors*, 22(21), 8090; https://doi.org/10.3390/s22218090

Sandler, M., Howard, A., Zhu, M., Zhmoginov, A., & Chen, L. C. (2018). MobileNetV2: Inverted Residuals and Linear Bottlenecks. CVPR. https://openaccess.thecvf.com/content_cvpr_2018/papers/Sandler_MobileNetV2_Inverted_Residuals_CVPR_2018_paper.pdf

Schwendeman, M. & Thomson, D. (2015). *A Horizon-Tracking Method for Shipboard Video Stabilization.* Journal of Atmospheric and Oceanic Technology, 32(1), 1–17.” https://doi.org/10.1175/JTECH-D-14-00047.1

Shao, Z., Wu, W., Wang, Z., Du, W. and Li, C. (2018). *SeaShips: A Large-Scale Precisely Annotated Dataset for Ship Detection. IEEE Access*, 6, 27780–27791. https://ieeexplore.ieee.org/document/8438999

Shedrawi, G., et al. (2024). Leveraging deep learning and computer vision technologies to enhance management of coastal fisheries in the Pacific region. *Scientific Reports*, 14(20915). https://doi.org/10.1038/s41598-024-71763-y

Simonyan, K., & Zisserman, A. (2015). *Very Deep Convolutional Networks for Large-Scale Image Recognition.* International Conference on Learning Representations (ICLR), https://www.robots.ox.ac.uk/~vgg/publications/2015/Simonyan15/simonyan15.pdf

Srivastava, N., Hinton, G., Krizhevsky, A., Sutskever, I., & Salakhutdinov, R. (2014). Dropout: A simple way to prevent neural networks from overfitting. *Journal of Machine Learning Research*, 15, 1929–1958. https://jmlr.org/papers/v15/srivastava14a.html

Talpur, K., Hasan, R., Gocer, I., Ahmad, S. and Bhuiyan, Z. (2025). AI in Maritime Security: Applications, Challenges, Future Directions, and Key Data Sources. *Information*, 16(8), 658. https://doi.org/10.3390/info16080658

Tan, M., & Le, Q. (2019). EfficientNet: Rethinking Model Scaling for Convolutional Neural Networks. Proceedings of the 36th International Conference on Machine Learning, ICML 2019, Long Beach, 9-15 June, 6105-6114. http://proceedings.mlr.press/v97/tan19a.html

Tian, Y., Ye, Q. and Doermann, D. (2025). YOLOv12: Attention-Centric Real-Time Object Detectors. https://arxiv.org/abs/2502.12524

**Ultralytics (2024).** *YOLOv8: Next-Generation Object Detection. Technical Documentation.* Retrieved from https://docs.ultralytics.com/models/yolov8/

Vaswani, A., Shazeer, N., Parmar, N., Uszkoreit, J., Jones, L., Gomez, A. N., Kaiser, L. and Polosukhin, I. (2017). *Attention is all you need.* Advances in Neural Information Processing Systems, 30, 5998–6008. https://arxiv.org/abs/1706.03762

Wang, C. Y., Yeh, I. H., & Liao, H. Y. M. (2024a). *YOLOv9: Learning what you want by controlling what you do not. arXiv preprint* arXiv:2402.13616.

Wang, C. Y. and Liao, H. Y. M. (2024b). *YOLOv10: Real-time object detection explained. https://arxiv.org/abs/2405.14458*.

Wang, L., Fan, S., Liu, Y., Li, Y., Fei, C., Liu, J., Liu, B., Dong, Y., Liu, Z. and Zhao, X. (2021). *A Review of Methods for Ship Detection with Electro-Optical Images in Marine Environments. Journal of Marine Science and Engineering*, 9(12), 1408; https://doi.org/10.3390/jmse9121408

Weiss, K., Khoshgoftaar, T., & Wang, D. (2016). A survey of transfer learning. *Journal of Big Data*, 3(9), 2-40. DOI:10.1186/s40537-016-0043-6

Zhu, F., Pan, B. and Im, N. (2025). *DMODT: Dynamic maritime environment object detection and tracking model for MASS. Ocean Engineering*, 340 (2), 122270. https://doi.org/10.1016/j.oceaneng.2025.122270